\pdfoutput=1

\documentclass[11pt]{article}

\usepackage[preprint]{acl}

\usepackage{times}
\usepackage{latexsym}
\usepackage{amsmath} 
\usepackage{mathbbol}
\usepackage{multirow}

\usepackage[T1]{fontenc}

\usepackage[utf8]{inputenc}

\usepackage{microtype}

\usepackage{inconsolata}

\usepackage{graphicx}

\title{COOL: Comprehensive Knowledge Enhanced Prompt Learning for Domain Adaptive Few-shot Fake News Detection}

\author{
 \textbf{Yi Ouyang\textsuperscript{1,2}},
 \textbf{Peng Wu\textsuperscript{1,2,\thanks{Corresponding Author.}}},
 \textbf{Li Pan\textsuperscript{1,2}}
\\
\\
 \textsuperscript{1}Shanghai Jiao Tong University, Shanghai, China
 \\
 \textsuperscript{2}Shanghai Key Laboratory of Integrated Administration Technologies for \\ Information Security, Shanghai Jiao Tong University, Shanghai, China
\\
\texttt{\{halaoyy, catking, panli\}@sjtu.edu.cn}
}

\begin{document}
\maketitle
\begin{abstract}
Most Fake News Detection (FND) methods often struggle with data scarcity for emerging news domain. Recently, prompt learning based on Pre-trained Language Models (PLM) has emerged as a promising approach in domain adaptive few-shot learning, since it greatly reduces the need for labeled data by bridging the gap between pre-training and downstream task. Furthermore, external knowledge is also helpful in verifying emerging news, as emerging news often involves timely knowledge that may not be contained in the PLM’s outdated prior knowledge. To this end, we propose COOL, a Comprehensive knOwledge enhanced prOmpt Learning method for domain adaptive few-shot FND. Specifically, we propose a comprehensive knowledge extraction module to extract both structured and unstructured knowledge that are positively or negatively correlated with news from external sources, and adopt an adversarial contrastive enhanced hybrid prompt learning strategy to model the domain-invariant news-knowledge interaction pattern for FND. Experimental results demonstrate the superiority of COOL over various state-of-the-arts.
\end{abstract}

\section{Introduction}

Emerging news domain with limited labeled data often have distinctive semantic characteristics other than historical news domain with sufficient labeled data, leading to degenerated performance for PLM-based FND methods which have to be fine-tuned on large-scale labeled data. To improve FND on emerging target domain, various domain adaptive fine-tuning strategies on PLM have been investigated \citep{mehta2022tackling, kaliyar2021fakebert, 10.1145/3581783.3612501, mridha2021comprehensive}. However, fine-tuning PLM is inherently data-intensive, as it requires additional supervised signals to adapt PLM from pre-training task to downstream task. Recently, prompt learning which bridges the gap between pre-training and downstream task by keeping downstream learning the same as the pre-training process, achieves success in few-shot scenarios and has been used in various domain adaptive tasks \citep{bai2024prompt, 10.1145/3589334.3645337, ge2023domain}.

\begin{figure}[t]
  \centering
  \includegraphics[width=\linewidth]{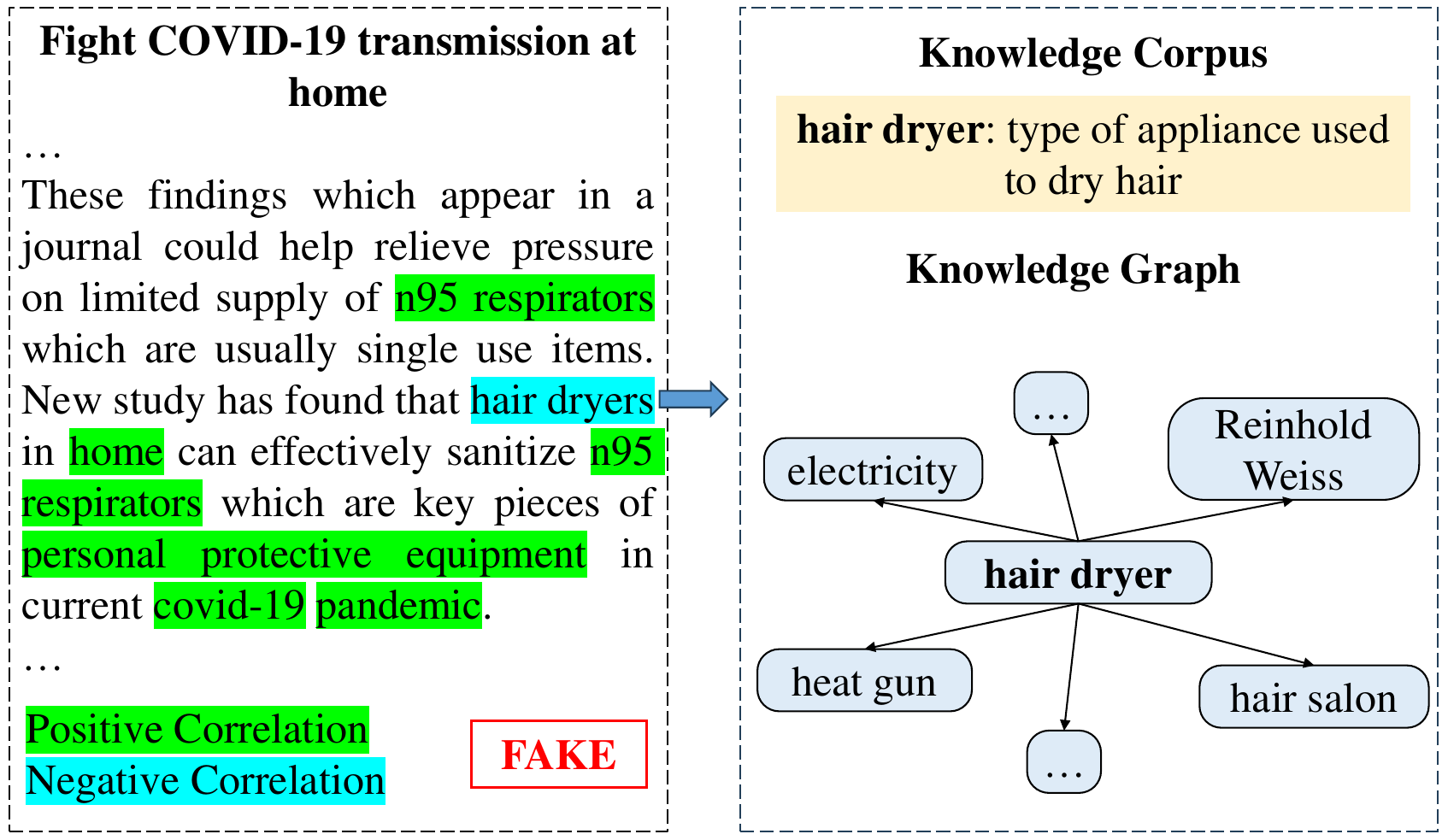}
  \caption{A piece of real-world fake news about the prevention of Covid-19.}
  \label{fig:motiv}
\end{figure}

Despite its promising performance, PLM's prior knowledge is constrained by the outdated pre-training corpus, leading to sub-optimal detection performance in emerging news domain where timely and domain-specific knowledge is involved. Therefore, it is crucial to leverage up-to-date heterogeneous external knowledge, including structured knowledge graph with relational knowledge among entities and unstructured knowledge corpus with descriptions about entity properties \citep{speer2017conceptnet, pei2023few}, to assist in domain adaptive FND. Most previous studies generally extract entity knowledge positively correlated with news \citep{dun2021kan, tseng2022kahan, ma2023kapalm}. However, the negatively correlated knowledge, i.e., entity knowledge not very correlated with news semantics, may also contributes considerably to FND. For example, Figure~\ref{fig:motiv} illustrates a fake news concerning the prevention of Covid-19, where exists entities positively correlated with news semantics like “pandemic”, as well as a negatively correlated entity “hair dryer”. Intuitively, the negatively correlated entity “hair dryer”, whose knowledge greatly deviates from the news semantics, significantly reveals the authenticity of the news. Hence both positively and negatively correlated knowledge in heterogeneous source should be comprehensively extracted for domain adaptive few-shot FND.

Existing knowledge enhanced FND models typically inject knowledge by concatenating it with the learned news features before the final classifier and adopting much labeled data to capture their interaction patterns \citep{ma2023kapalm, dun2021kan}. Such scheme may not be applicable to few-shot prompt learning, as its input feature for classifier is the learned embedding of a mask word where the knowledge-news interaction patterns are hard to be captured. Another intuitive scheme is to incorporate the knowledge into the prompt template before PLM encoder. However, both hand-crafted and soft prompt template may not be suitable for directly injecting knowledge, as hard template cannot flexibly inject various forms and quantities of knowledge, while soft template may struggle to fit the FND task. As a result, hybrid templates have been adopted, which incorporate knowledge representation into several soft prompt vectors, while guide PLM in reasoning about news authenticity via hard templates \citep{jiang2022fake}. Despite their effectiveness in modeling relationships between news, knowledge and detection task, their performances can be further improved in domain adaptive few-shot scenario by capturing the domain-invariant interaction patterns.

To this end, we propose COOL for domain adaptive few-shot FND, which extracts comprehensive knowledge that are positively or negatively correlated with news from heterogeneous sources, and injects it into prompt learning by an adversarial contrastive enhanced hybrid prompt learning framework. More specifically, a comprehensive knowledge extraction module is proposed to retrieve both structured relational and unstructured descriptive knowledge from external sources, and filter both positively and negatively correlated knowledge via a signed correlation-aware attention. The filtered comprehensive knowledge is incorporated by a hybrid prompt learning framework, where prefix soft prompt composed of several learnable tokens is used to receive knowledge representations flexibly, while postfix hand-crafted hard prompt facilitates PLM modeling task-specific interaction between knowledge and news. The adversarial contrastive learning is applied to facilitate the model capturing domain-invariant news-knowledge interaction patterns to improve the domain adaptive few-shot FND performance. Experimental studies validate the benefits of incorporating comprehensive knowledge into prompt learning for domain adaptive few-shot FND. The primary contributions of this paper can be summarized as:

(1) We highlight that the comprehensive knowledge positively or negatively correlated with news is crucial for PLM to detect fake news in emerging domains, which can be extracted from heterogeneous source.

(2) We propose COOL, which devises a comprehensive knowledge extraction module to extract knowledge and injects it into a hybrid prompt learning framework to model domain-invariant news-knowledge interaction patterns.

(3) Experiments on real-word datasets are conducted to demonstrate that COOL consistently outperforms the several state-of-the-arts.


\section{Related Works}

\textbf{Domain Adaptive Few-shot News Detection}. Previous FND methods focus on modeling fake news patterns by PLM-based models fine-tuned on large-scale labeled datasets \citep{mehta2022tackling, kaliyar2021fakebert, 10.1145/3589334.3645468, mridha2021comprehensive}. However, it is frequent to face the data scarcity issue of emerging news domain. To tackle this problem, many domain adaptive few-shot FND methods have adopted various techniques to adapt the domain-invariant features learned from the abundant source domain data to the target news domain with limited labeled data \citep{yue2022contrastive, mosallanezhad2022domain, lin2022detect, ran2023unsupervised}, such as meta-learning  improving domain adaptation by adjusting model parameters step by step across tasks \citep{yue2023metaadapt, nan2022improving, hospedales2021meta}, and contrastive learning reducing the inter-domain discrepancy by appropriate contrastive loss \citep{yue2022contrastive, lin2022detect, ran2023unsupervised}. More recently, prompt learning, which bridges the gap between PLM's pre-training and downstream task, exhibits significant successes in many domain adaptive few-shot tasks, such as rumor detection \citep{lin2023zero} and dialogue summarization \citep{zhao2022adpl}. Despite its success in various tasks, its performance may be constrained in domain adaptive FND, as the emerging news typically involves timely and domain-specific knowledge that may not be included in PLM’s outdated pre-training corpus. This inspires us to design a knowledge enhanced prompt learning method for better domain adaptive few-shot FND.

\textbf{Knowledge Enhanced Fake News Detection}. News naturally encompasses a number of knowledge entities, whose knowledge can serve as critical evidence for news verification, inspiring researchers to investigate knowledge-enhanced methods for FND \citep{dun2021kan, tseng2022kahan, ma2023kapalm, zhang2024reinforced}. Most existing methods leverage structured knowledge graph, e.g. ConceptNet \citep{speer2017conceptnet} and YAGO \citep{suchanek2007yago}, to capture the relational knowledge among entities for FND \citep{ma2023kapalm, kim2023factkg, sun2023inconsistent}. Some studies benefit from abundant heterogeneous knowledge by exploiting both structured knowledge graph and unstructured knowledge corpus, e.g., Wikipedia corpus \citep{hu2021compare, zou2023decker}. Timely and rich external knowledge can compensate for the knowledge gap of PLM in emerging news, thereby improving the domain adaptive few-shot FND performance of PLM based methods, including prompt learning. KPL \citep{jiang2022fake} devises a knowledgeable prompt learning framework which incorporates the sequential knowledge entities into prompt template to predict the news veracity. Different from \citet{jiang2022fake}, the proposed COOL extracts more comprehensive knowledge that have either positive or negative correlation with news from both structured and unstructured external knowledge sources, which is further incorporated into an adversarial contrastive enhanced hybrid prompt learning framework to model the domain-invariant interaction patterns between news and knowledge for domain adaptive few-shot FND.

\section{Problem Statement}
Let $\mathcal{D}_s=\left\{\left(\mathcal{X}_1^s, y_1^s\right),\left(\mathcal{X}_2^s, y_2^s\right), \ldots,\left(x_M^s, y_M^s\right)\right\}$ and $\mathcal{D}_t=\left\{\left(\mathcal{X}_1^t, y_1^t\right),\left(\mathcal{X}_2^t, y_2^t\right), \ldots,\left(x_N^t, y_N^t\right)\right\}$ denote the datasets of source and target domain, respectively, where $M$, $N$ denote the numbers of news in source and target domain, respectively. Each news $\mathcal{X}=\left\{w_i\right\}$ consists of a sequence of words. The label $y \in[0,1]$ denotes the veracity of news, where 0 indicates true and 1 indicates fake. The domain adaptive few-shot FND is defined as: given the source domain dataset $\mathcal{D}_s$ and limited access to the target domain dataset, i.e., only a K-shot subset $\mathcal{D}_t^{\prime} \subset \mathcal{D}_t$ is available for training where $K \ll N$, the goal is to correctly predict the veracity of news in the target domain dataset $\mathcal{D}_t$.

\begin{figure*}[!tb!]
  \centerline{\includegraphics[width=0.85\textwidth]{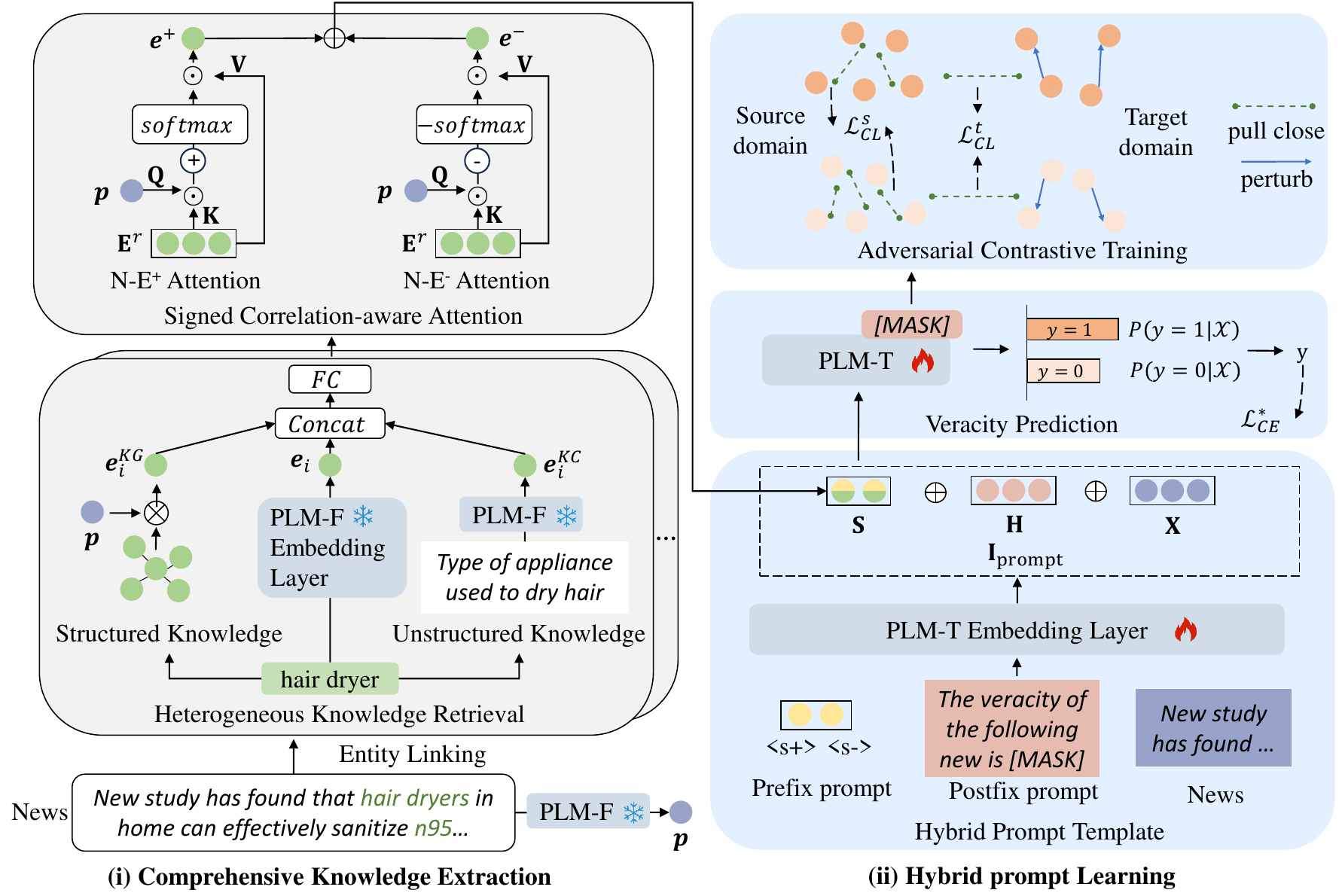}}
  \caption{Architecture of the proposed COOL model.}
  \label{fig:cool}
\end{figure*}

\section{Methodologies}
The architecture of the proposed model is illustrated in Figure~\ref{fig:cool}. It consists of two modules: (i) comprehensive knowledge extraction, which extracts structured and unstructured knowledge from external sources that are positively or negatively correlated with news; (ii) hybrid prompt learning, in which a hybrid prompt template is devised to simultaneously incorporate external knowledge and guide PLM on FND task, and an adversarial contrastive training strategy is leveraged to capture the domain-invariant news-knowledge interaction pattern. Each of module is described in details next.

\subsection{Comprehensive Knowledge Extraction}

\subsubsection{Heterogeneous Knowledge Retrieval}
To retrieve external knowledge critical for verifying news, we first identify the knowledge entities $\mathcal{E}=\left\{e n_i\right\}$ from a given news $\mathcal{X}$ by an entity linking method \citep{ferragina2010tagme}. The identified entities are embedded by the embedding layer of a parameter-frozen PLM, i.e., $\mathbf{E}=\left[\boldsymbol{e}_1, \ldots, \boldsymbol{e}_{|\mathcal{E}|}\right] \in \mathbb{R}^{|\mathcal{E}| \times d}$, where $\boldsymbol{e}_i$ is the averaged word embedding for entity $en_i$ and $d$ is the hidden dimension of PLM. The parameter-frozen PLM, denoted as PLM-F, is used to represent various information and will not attend training. The knowledge entities usually associate with heterogeneous external knowledge, including structured relational knowledge from knowledge graph, and unstructured descriptive knowledge from knowledge corpus. Therefore, we propose a structured knowledge retriever and an unstructured knowledge retriever for heterogeneous knowledge.

\textbf{Structured Knowledge Retriever}. Knowledge graph $\mathcal{G}$ encompasses structured knowledge in form of triples $\left(e n_s, r e l, e n_t\right)$, where $rel$ is the relation between two entities. The structured knowledge of an entity $en_i$ is set as its neighbors $\mathcal{N}\left(e n_i\right)=\left\{e n \mid\left(e n_i, r e l, e n\right) \in \mathcal{G} \vee \left(e n, r e l, e n_i\right) \in \mathcal{G}\right\}$. Since a news entity may have many neighbors with various semantics, not all of its neighbors contribute equally for verifying a news. For instance, when verifying a news reporting Donald Trump signed the Iran Deal, the structured knowledge \textit{(Donald Trump, significant event, United States withdrawal from Iran Deal)} is more informative than other knowledge of entity "Donald Trump". Hence the structured knowledge of each entity should be filtered based on their relevance with news semantics. The typical relevant information filter method is attention mechanism \citep{vaswani2017attention}. However, since attention mechanism will also be used to filter relevant entities in news latter, if it is used simultaneously to filter the structured knowledge of each entity, the nested attention will be formed and lead to exponential increase in computational complexity. Previous works abandon the attentive filtering of structured knowledge by leveraging \emph{mean pooling} to avoid the nested attention\citep{dun2021kan, tseng2022kahan}. Instead, we adopt a modulation mechanism based on \emph{product \& max pooling} \citep{ouyang2022asymmetrical} to attentively filter the related structured knowledge with lower computational complexity. Specifically, given an entity $e n_i$ and its neighbors $\mathcal{N}\left(e n_i\right)$, the structured knowledge is filtered as:
\begin{equation}
\boldsymbol{e}_i^{K G}=M P_{e n \in \mathcal{N}\left(e n_i\right)}\left(\boldsymbol{p} \otimes \boldsymbol{e}_{e n}\right)
\end{equation}
where $\boldsymbol{p}$ is the news representation embedded by PLM-F on its content, $MP$ indicates the max pooling operation, $\otimes$ stands for the element-wise product, and $\boldsymbol{e}_{en}$ is the embedding of a neighbor entity. The element-wise product evaluates the relatedness between each neighbor and news, while the max pooling helps to focus on the most related knowledge from neighbors and reduce noises.

\textbf{Unstructured Knowledge Retriever}. The unstructured descriptive knowledge from knowledge corpus is another important supplement for FND, since it describes connotation and properties of each entity with natural language, whose semantics can also interact with news. For example, a short description “contagious disease caused by SARS-CoV-2” of entity “Covid-19” can interact with news by offering PLM with knowledge lacked during pre-training. The unstructured knowledge $\boldsymbol{e}_i^{K C} \in \mathbb{R}^d$ of a given entity $e n_i$ can also be embedded by applying PLM-F on its description sentence.

Heterogeneous knowledge complements each other for FND. The final knowledge $\boldsymbol{e}_i^{r} \in \mathbb{R}^d$ of an entity $en_i$ is extracted by concatenating its structured knowledge $\boldsymbol{e}_i^{KG}$ , initial entity embedding $\boldsymbol{e}_{i}$ and unstructured knowledge $\boldsymbol{e}_i^{KC}$ and passing through a fully connected layer $FC$:
\begin{equation}
\boldsymbol{e}_i^r=F C\left(\left[\boldsymbol{e}_i^{K G} ; \boldsymbol{e}_i ; \boldsymbol{e}_i^{K C}\right]\right)
\end{equation}

\subsubsection{Signed Correlation-aware Attention}
The knowledge either positively or negatively correlated with news is significant for FND, as the positive one provides news-related knowledge context, while the negative one reveals news-knowledge discrepancy \citep{dun2021kan, sun2023inconsistent}. To capture both positively and negatively correlated knowledge, a signed correlation-aware attention consisting of \emph{News towards Positively correlated Entity} (N-$\mathrm{E^{+}}$) attention and \emph{News towards Negatively correlated Entity} (N-$\mathrm{E^{-}}$) attention is devised.

\textbf{N-$\mathrm{\mathbf{E}^{+}}$ Attention}. N-$\mathrm{E^{+}}$ attention follows typical attention mechanism which captures positively correlated knowledge by assigning greater importance for entity knowledge that is more correlated with news semantics:
\begin{equation}
Attn^{+}(\mathbf{Q}, \mathbf{K}, \mathbf{V})=\operatorname{softmax}\left(\frac{\mathbf{Q} \mathbf{K}^T}{\sqrt{d_K}}\right) \mathbf{V}
\end{equation}
where $d_K$ is the dimension of keys. The dot-product attention function measures the positive correlation between queries and keys and assigns weights attentively to values. Hence the positively correlated knowledge $\boldsymbol{e}^{+}$ can be extracted by setting the news representation $\boldsymbol{p}$ as queries, and the all extracted knowledge $\mathbf{E}^r=\left[\boldsymbol{e}_1^r, \ldots, \boldsymbol{e}_{|\mathcal{E}|}^r\right] \in \mathbb{R}^{|\mathcal{E}| \times d}$ as keys and values:
\begin{equation}
\boldsymbol{e}^{+}=Attn^{+}\left(\boldsymbol{p} \mathbf{W}_Q^{+}, \mathbf{E}^r \mathbf{W}_K^{+}, \mathbf{E}^r \mathbf{W}_V^{+}\right)
\end{equation}
where $\mathbf{W}_Q^{+}$, $\mathbf{W}_K^{+}$, $\mathbf{W}_V^{+}$ are learnable parameters.

\textbf{N-$\mathrm{\mathbf{E}^{-}}$ Attention}. (N-$\mathrm{E^{-}}$) attention captures negatively correlated knowledge by assigning greater importance for entity knowledge that is less correlated with news semantics:
\begin{equation}
Attn^{-}(\mathbf{Q}, \mathbf{K}, \mathbf{V})=-\operatorname{softmax}\left(-\frac{\mathbf{Q} \mathbf{K}^T}{\sqrt{d_K}}\right) \mathbf{V}
\end{equation}
The minus inside the dot-product attention function assigns greater weights to keys that are less correlated with queries, and the outside minus further reverses the direction of the resulted feature vector to distance it from the result of N-$\mathrm{E^{+}}$ attention. Similarly, the negatively correlated knowledge $\boldsymbol{e}^{-}$ is extracted as:
\begin{equation}
\boldsymbol{e}^{-}=Attn^{-}\left(\boldsymbol{p} \mathbf{W}_Q^{-}, \mathbf{E}^r \mathbf{W}_K^{-}, \mathbf{E}^r \mathbf{W}_V^{-}\right)
\end{equation}
where $\mathbf{W}_Q^{-}$, $\mathbf{W}_K^{-}$, $\mathbf{W}_V^{-}$ are learnable parameters. 

Finally, we obtain the comprehensive knowledge as $\mathbf{E}^c=\left[\boldsymbol{e}^{+}, \boldsymbol{e}^{-}\right] \in \mathbb{R}^{2 \times d}$ for FND.

\subsection{Hybrid Prompt Learning}

\subsubsection{Hybrid Prompt Template}
To inject comprehensive knowledge into prompt learning, a hybrid prompt template consists of both prefix soft prompt and postfix hard prompt is adopted. The soft prompt composed of several learnable tokens receive comprehensive knowledge freely by generating appropriate semantics, while the hard prompt is a manually designed natural language sentence used to guide PLM in reasoning about news authenticity. Specifically, two tokens <s+>, <s-> with randomly initialized learnable embeddings $\left[\boldsymbol{s}^{+}, \boldsymbol{s}^{-}\right] \in \mathbb{R}^{2 \times d}$ are set as prefix soft prompt to receive the positively and negatively correlated knowledge, respectively. The soft prompt embedding after receiving knowledge is:
\begin{equation}
\mathbf{S}=\frac{1}{2}\left(\left[\boldsymbol{s}^{+},\boldsymbol{s}^{-}\right]+\mathbf{E}^c\right)
\end{equation}
While for postfix hard prompt, we test and adopt a cloze-style natural language sentence specialized for FND task, e.g., \emph{“The veracity of the following news is [MASK].”}. The hard prompt embedding is denoted as $\mathbf{H}=\left[\boldsymbol{h}_1, \ldots, \boldsymbol{h}_{[\mathrm{MASK}]}, \ldots, \boldsymbol{h}_{n_h}\right] \in \mathbb{R}^{n_{h} \times d}$ where $n_{h}$ is the number of hard prompt tokens, and $\boldsymbol{h}_{[\mathrm{MASK}]}$ is the embedding of [MASK] token. It is got from the embedding layer of a tunable PLM called PLM-T.

The token embeddings of the given news $\mathcal{X}$, i.e., $\mathbf{X}=\left[\boldsymbol{x}_1, \ldots, \boldsymbol{x}_{n_x}\right] \in \mathbb{R}^{n_x \times d}$, is also got from the embedding layer of PLM-T, where $n_x$ is the number of news tokens. The final hybrid prompt template $\mathbf{I}_{\mathrm{prompt}}$ is:
\begin{equation}
\mathbf{I}_{\mathrm{prompt }}=\left[ \mathbf{S}, \mathbf{H}, \mathbf{X}\right]
\end{equation}

\subsubsection{Veracity Prediction}
$\mathbf{I}_{\mathrm{prompt}}$ is then sent to transformer layers of PLM-T to predict news veracity. Specifically, the output embedding $\boldsymbol{o}_{[\mathrm{MASK] }} \in \mathbb{R}^d$ of [MASK] token is obtained as:
\begin{equation}
\boldsymbol{o}_{\mathrm{[MASK] }}=\mathrm{PLM}\text{-}\mathrm{T}\left(\mathbf{I}_{\mathrm {prompt }}\right)
\end{equation}
Its vocabulary distribution $\boldsymbol{v}_{[\mathrm{MASK] }} \in \mathbb{R}^{|\mathcal{V}|}$ is got by sending $\boldsymbol{o}_{\mathrm{[MASK] }}$ to the head function of PLM, where $\mathcal{V}$ is the vocabulary of PLM. We manually define the vocabulary subsets $\mathcal{V}_*=\left\{\mathcal{V}_0, \mathcal{V}_1\right\}$, where $\mathcal{V}_0$ contains words about true, $\mathcal{V}_1$ contains words about fake. Then the probability of each label $y \in[0,1]$ for a given news $\mathcal{X}$ is calculated as:
\begin{equation}
P(y \mid \mathcal{X})=\frac{\exp \left(\boldsymbol{v}_{[\mathrm{MASK] }}\left(\mathcal{V}_y\right)\right)}{\sum_{\mathcal{V}_i \in \mathcal{V}_*} \exp \left(\boldsymbol{v}_{[\mathrm {MASK] }}\left(\mathcal{V}_i\right)\right)}
\end{equation}
where $\boldsymbol{v}_{[\mathrm {MASK] }}\left(\mathcal{V}_i\right)$ is sum of distribution scores of words in $\mathcal{V}_i$. Finally, the cross-entropy loss in each domain is as below, where $* \in\{s, t\}$ and $\mathcal{D}_*$ is the source or target domain dataset:
\begin{equation}
\begin{split}
    \mathcal{L}_{C E}^*=-\sum_{\left(\mathcal{X}_i^*,y_i^*\right) \in\mathcal{D}_*} y_i^*\log\left(P\left(y_i^*\mid \mathcal{X}_i^*\right)\right)
\end{split}
\end{equation}

\subsubsection{Adversarial Contrastive Training}
In domain adaptive few-shot scenario, the detection performance is inherently determined by the quantity and quality of target domain samples, which are limited and may suffer from noises. To alleviate this problem, adversarial samples are generated by adding each target domain sample with a worst-case perturbation, i.e., a normed noisy vector towards the gradient direction that maximizes the loss $\mathcal{L}_{C E}^t$: 
\begin{equation}
\boldsymbol{o}_{[\mathrm{MASK}], \mathrm{adv}}^t=\boldsymbol{o}_{[\mathrm{MASK}]}^t+\frac{\nabla \mathcal{L}_{C E}^t}{\left\|\nabla \mathcal{L}_{C E}^t\right\|}
\end{equation}
where $\nabla \mathcal{L}_{C E}^t$ is the first-order gradient of $\mathcal{L}_{C E}^t$, which is approximated by the Fast Gradient Value \citep{rozsa2016adversarial} method. The adversarial samples serve as additional target domain samples in training.

To facilitate PLM modeling the domain-invariant news-knowledge interaction pattern, we adopt contrastive loss function to reduce the inter-domain discrepancy by explicitly pulling close the output [MASK] embedding of news with the same label from target and source domains, respectively: 
\begin{equation}  
\begin{split}  
\mathcal{L}_{C L}^t &= -\frac{1}{N \times M} \sum_{\left(\mathcal{X}_i^t, y_i^t\right) \in \mathcal{D}_t} \sum_{\left(\mathcal{X}_j^s, y_j^s\right) \in \mathcal{D}_s}  \mathbb{1}_{\left[y_j^t=y_i^s\right]} \\
& \log \frac{\exp \left(S\left(\boldsymbol{o}_i^t, \boldsymbol{o}_j^s\right) / \tau\right)}{\sum_{\left(\mathcal{X}_k^s, y_k^s\right) \in \mathcal{D}_s} \exp \left(S\left(\boldsymbol{o}_i^t, \boldsymbol{o}_k^s\right) / \tau\right)}  
\end{split}  
\end{equation} 
where $S(\cdot)$ is cosine similarity, $\tau$ is a temperature parameter. Similarly, another contrastive loss is utilized to reduce the intra-class discrepancy for abundant source domain samples:
\begin{equation} 
\begin{split}  
\mathcal{L}_{C L}^s &= -\frac{1}{M \times(M-1)} \\  
&\sum_{\left(\mathcal{X}_i^s, y_i^s\right) \in \mathcal{D}_s} \sum_{\left(\mathcal{X}_j^s, y_j^s\right) \in \mathcal{D}_s} \mathbb{1}_{[i \neq j]} \mathbb{1}_{\left[y_i^s=y_j^s\right]} \\  
&\log \frac{\exp \left(S\left(\boldsymbol{o}_i^s, \boldsymbol{o}_j^s\right) / \tau\right)}{\sum_{\left(\mathcal{X}_k^s, y_k^s\right) \in \mathcal{D}_s} \mathbb{1}_{[i \neq k]} \exp \left(S\left(\boldsymbol{o}_i^s, \boldsymbol{o}_k^s\right) / \tau\right)}  
\end{split}  
\end{equation}

The final loss of our model is then formulated as below, where $\alpha$ is a trade-off parameter:
\begin{equation}
\mathcal{L}=\sum_{* \in\{s, t\}} \alpha \mathcal{L}_{C E}^*+(1-\alpha) \mathcal{L}_{C L}^*
\end{equation}

\section{Experiments}
\subsection{Experiment Setup}
\textbf{Dataset}. Three datasets are utilized to implement the experiments in domain adaptive few-shot setting. Snopes \citep{popat2017truth} is a domain-agnostic dataset providing news in various domains and is adopted as source domain dataset. Politifact \citep{shu2020fakenewsnet} is a dataset specialized for US political system. CoAID \citep{cui2020coaid} is a healthcare dataset containing COVID-19 related news. They are domain-specific datasets used as target domain datasets. The statistics of the datasets are reported in Table~\ref{tab: statis} in Appendix \ref{sec:data_stat}.

\begin{table*}
\centering
\caption{Comparisons of different models on domain adaptive few-shot FND task. The best results are in boldface and the second-best results are underlined.}
\label{tab: comp}
\resizebox{\textwidth}{33mm}{
\begin{tabular}{c|c|c|cccccccccc} 
\hline
Target
  (Source)                      & \# Shot             & Metric        & TextCNN & Bi-LSTM & KAN    & FT & ACLR           & PET    & Soft-PT & RPL            & KPL            & COOL             \\ 
\hline
\multirow{8}{*}{Politifact
  (Snopes)} & \multirow{2}{*}{2}  & \textit{Acc.} & 0.5730  & 0.5730  & 0.5732 & 0.6171     & \underline{0.6388} & 0.6208 & 0.5400  & 0.6316         & 0.6232         & \textbf{0.6575}  \\
                                       &                     & \textit{F1}   & 0.5728  & 0.5710  & 0.5669 & 0.5947     & \underline{0.6375} & 0.5982 & 0.5180  & 0.6196         & 0.6088         & \textbf{0.6505}  \\ 
\cline{2-13}
                                       & \multirow{2}{*}{4}  & \textit{Acc.} & 0.6070  & 0.6093  & 0.5741 & 0.6350     & 0.6578         & 0.6660 & 0.5698  & \underline{0.6828} & 0.6630         & \textbf{0.7010}  \\
                                       &                     & \textit{F1}   & 0.5952  & 0.5965  & 0.5674 & 0.6210     & 0.6550         & 0.6312 & 0.5454  & \underline{0.6762} & 0.6487         & \textbf{0.6843}  \\ 
\cline{2-13}
                                       & \multirow{2}{*}{8}  & \textit{Acc.} & 0.6348  & 0.6490  & 0.6155 & 0.6691     & \underline{0.7849} & 0.6793 & 0.6443  & 0.7742         & 0.7165         & \textbf{0.7869}  \\
                                       &                     & \textit{F1}   & 0.6060  & 0.6202  & 0.6117 & 0.6683     & \underline{0.7243} & 0.6754 & 0.6326  & 0.7072         & 0.7100         & \textbf{0.7767}  \\ 
\cline{2-13}
                                       & \multirow{2}{*}{16} & \textit{Acc.} & 0.6674  & 0.6963  & 0.6229 & 0.7556     & 0.7882         & 0.7775 & 0.6892  & \underline{0.8394} & 0.8129         & \textbf{0.8430}  \\
                                       &                     & \textit{F1}   & 0.6587  & 0.6932  & 0.6222 & 0.7381     & 0.7776         & 0.7764 & 0.6882  & \underline{0.8296} & 0.8084         & \textbf{0.8329}  \\ 
\hline
\multirow{8}{*}{CoAID (Snopes)}        & \multirow{2}{*}{2}  & \textit{Acc.} & 0.3522  & 0.4129  & 0.4438 & 0.4626     & 0.5323         & 0.4315 & 0.4882  & 0.5433 & \underline{0.5494} & \textbf{0.6022}  \\
                                       &                     & \textit{F1}   & 0.2879  & 0.3284  & 0.3538 & 0.3615     & 0.4015         & 0.3616 & 0.3620  & \underline{0.4169} & 0.3887         & \textbf{0.4470}  \\ 
\cline{2-13}
                                       & \multirow{2}{*}{4}  & \textit{Acc.} & 0.4009  & 0.4915  & 0.4845 & 0.4927     & 0.5527         & 0.5960 & 0.5376  & 0.6916         & \underline{0.7261} & \textbf{0.7316}  \\
                                       &                     & \textit{F1}   & 0.3239  & 0.3765  & 0.3780 & 0.3930     & 0.4341         & 0.4494 & 0.3988  & 0.5086         & \underline{0.5334} & \textbf{0.5513}  \\ 
\cline{2-13}
                                       & \multirow{2}{*}{8}  & \textit{Acc.} & 0.4647  & 0.5363  & 0.5335 & 0.5447     & 0.5997         & 0.6607 & 0.5944  & 0.7227         & \underline{0.7392} & \textbf{0.7409}  \\
                                       &                     & \textit{F1}   & 0.3695  & 0.4230  & 0.4142 & 0.4221     & 0.4654         & 0.4818 & 0.4002  & 0.5307         & \underline{0.5534} & \textbf{0.5578}  \\ 
\cline{2-13}
                                       & \multirow{2}{*}{16} & \textit{Acc.} & 0.4991  & 0.5470  & 0.5754 & 0.6474     & 0.6341         & 0.7136 & 0.6332  & 0.7336         & \underline{0.7562} & \textbf{0.7937}  \\
                                       &                     & \textit{F1}   & 0.3856  & 0.4371  & 0.4399 & 0.4697     & 0.4815         & 0.5328 & 0.4692  & 0.5542         & \underline{0.5609} & \textbf{0.5900}  \\
\hline
\end{tabular}}
\end{table*}

\textbf{Baseline}. The COOL is compared with several groups of models suitable for domain adaptive few-shot FND, which include neural network-based models: \textbf{TextCNN} \citep{chen2015convolutional} and \textbf{Bi-LSTM} \cite{bahad2019fake}; knowledge enhanced neural network-based model \textbf{KAN} \citep{dun2021kan}; PLM-based models: \textbf{FT} \citep{liu2019roberta}, \textbf{ACLR} \citep{lin2022detect}, \textbf{PET} \citep{schick2021exploiting}, \textbf{Soft-PT} \citep{li2021prefix} and \textbf{RPL} \citep{lin2023zero}; knowledge enhanced PLM-based model \textbf{KPL} \citep{jiang2022fake}. The baseline methods are elaborated in Appendix \ref{sec: baselines}.

\textbf{Implementation Details}. We use Pytorch to implement our model \footnote{The code will be released upon publication.}. For domain adaptive few-shot FND, the source domain dataset and a randomly selected $K$-shot subset of target domain dataset are available for model training, where $K \in\{2,4,8,16\}$. The rest part of target domain dataset is used as test set to evaluate the detection performance. \emph{Acc.} (Accuracy) and \emph{F1} (Macro F1 score) are adopted for evaluating the performance, which have been widely used in previous works \citep{zhang2024reinforced, lin2023zero}. More details of implementation can be found in Appendix \ref{sec: imple}.

\subsection{Main Results}
The comparison results are reported in Table~\ref{tab: comp}. It is shown that our proposed COOL consistently achieves the best performance in all settings, with average improvements of 2.14\% and 4.16\% compared to the second-best method on Politifact and CoAID, respectively. Specifically, COOL performs better than all PLM-based methods, confirming the effectiveness of external knowledge in improving PLM in domain adaptive few-shot FND. The superiority of COOL over KPL, another knowledge enhanced prompt learning method, is possibly because: (1) our model incorporates more comprehensive knowledge by extracting both structured and unstructured knowledge that are positively or negatively correlated with news. (2) we design an adversarial contrastive enhanced hybrid prompt learning framework which incorporates comprehensive knowledge flexibly with appropriately learned semantics and guides PLM in modeling domain-invariant news-knowledge interaction patterns. The detailed comparisons between baselines are discussed in Appendix \ref{sec: comp}.

\begin{figure}[h]
  \centering
  \includegraphics[width=\linewidth]{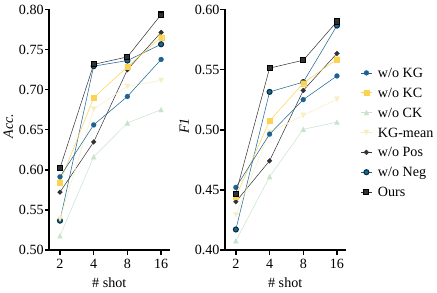}
  
  \caption{Ablation studies in comprehensive knowledge extraction module on CoAID, where “w/o KG” and “w/o KC” means removing structured knowledge retriever and unstructured knowledge retriever, respectively. “w/o CK” means removing the entire comprehensive knowledge extraction module. "KG-mean" means replacing the modulation mechanism with \emph{mean pooling} in structured knowledge retriever. “w/o Pos” and “w/o Neg” means removing N-$\mathrm{E^{+}}$ and N-$\mathrm{E^{-}}$ attention, respectively.}
  \label{fig: abla_know}
\end{figure}

\subsection{Abaltion Study}
Ablation studies are conducted to analyze the effects of the key designs in our model. The experimental results of ablation studies in comprehensive knowledge extraction module on CoAID are reported in Figure~\ref{fig: abla_know}. If the structured knowledge retriever is removed, our model drops averagely 6.46\% \emph{Acc.} and 5.58\% \emph{F1}. When the unstructured knowledge retriever is removed, it drops averagely 4.71\% \emph{Acc.} and 4.30\% \emph{F1}. If the entire comprehensive knowledge extraction module is eliminated, our model reduces averagely by 12.85\% \emph{Acc.} and 12.40\% \emph{F1}. These results validate that both structured relational knowledge and unstructured descriptive knowledge are helpful for FND and they complement with each other to provide comprehensive knowledge. Moreover, if the modulation mechanism in structured knowledge retriever is replaced by \emph{mean pooling}, COOL decreases averagely by 8.37\% \emph{Acc.} and 8.20\% \emph{F1}, which validates the effectiveness of modulation in attentively extracting structured knowledge and reducing noises. The efficacy of signed correlation-aware attention is further validated. Specifically, The model without N-$\mathrm{E^{+}}$ attention drops averagely 6.91\% \emph{Acc.} and 6.12\% \emph{F1}, while the model without N-$\mathrm{E^{-}}$ attention drops averagely 5.63\% \emph{Acc.} and 3.52\% \emph{F1}. This confirms that the knowledge either positively or negatively correlated with news provides critical evidences to verify news authenticity.

\begin{figure}[t]
  \centering
  \includegraphics[width=\linewidth]{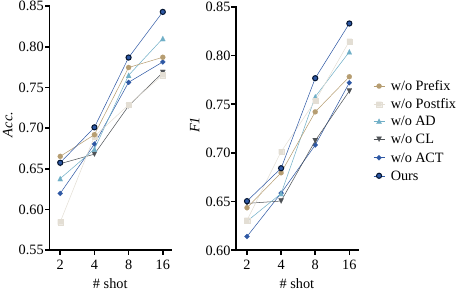}
  \caption{Ablation studies in hybrid prompt learning module on Politifact, where “w/o Prefix” and “w/o Postfix” means removing the prefix soft prompt and the postfix hard prompt, respectively. “w/o AD” and “w/o CL” means removing adversarial augmentation and contrastive training, respectively. “w/o ACT” means eliminates the adversarial contrastive training strategy.}
  \label{fig: abla_prompt}
\end{figure}



\begin{figure}[t]
  \centering
  \includegraphics[width=\linewidth]{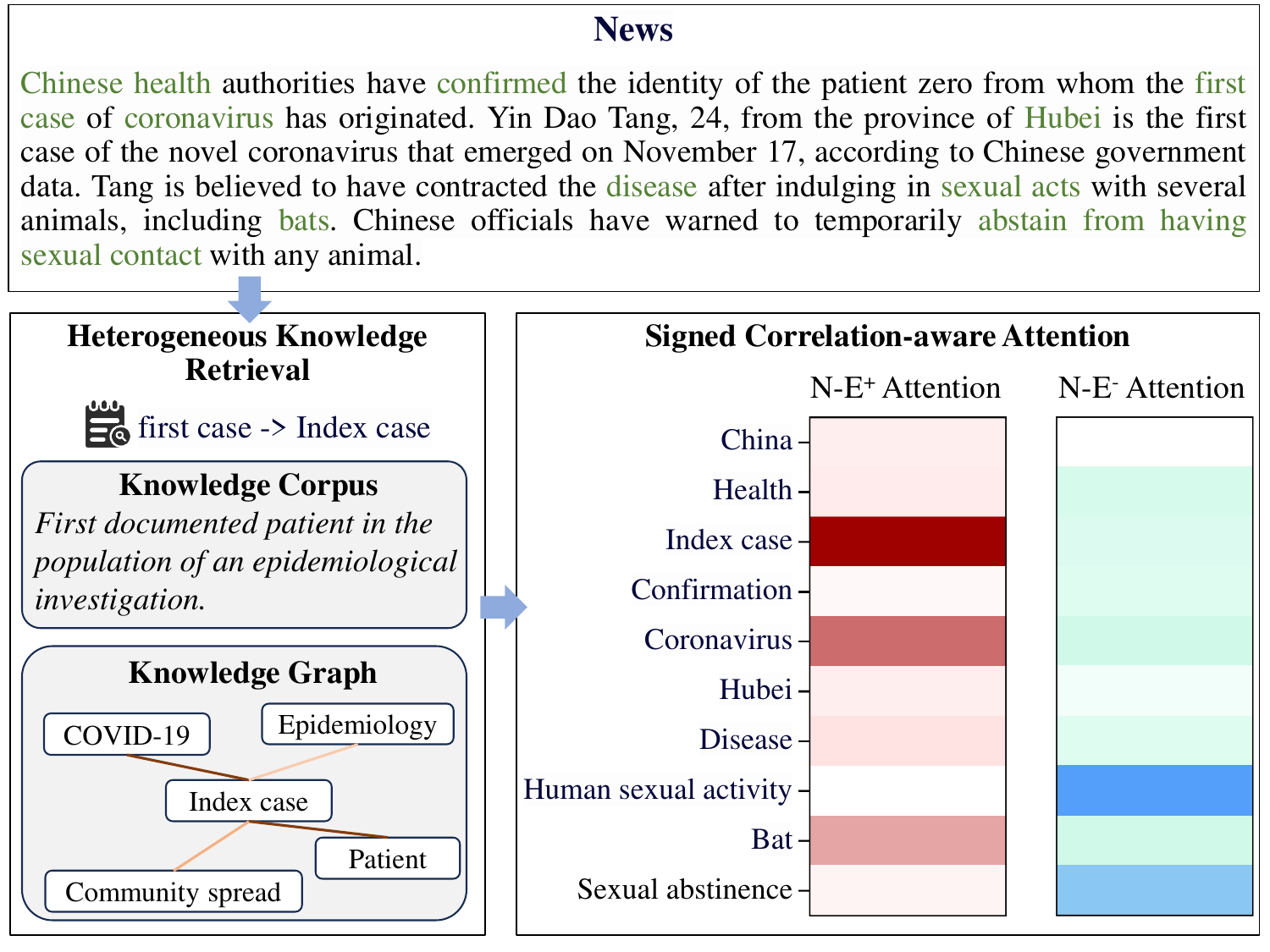}
  \caption{A real-world case from CoAID showing how COOL extracts comprehensive knowledge from external sources.}
  \label{fig: case}
\end{figure}

We also implement ablation experiments in hybrid prompt learning module on Politifact and report the results in Figure~\ref{fig: abla_prompt}. When the prefix prompt is eliminated and the knowledge is directly concatenated with postfix hard prompt, the model drops averagely 4.71\% and 3.18\% on \emph{Acc.} and \emph{F1}, which proves the advantage of incorporating knowledge with soft prompts to generate appropriate semantics. When the postfix hand-crafted prompt is removed, the model reduces averagely by 8.50\% \emph{Acc.} and 1.43\% \emph{F1}. This confirms the effect of task-specific hand-crafted prompt in guiding PLM in reasoning about news authenticity. The model removing adversarial augmentation drops averagely 3.36\% \emph{Acc.} and 3.22\% \emph{F1}, while the model removing contrastive training decreases averagely by 5.31\% \emph{Acc.} and 5.46\% \emph{F1}. If the entire adversarial contrastive training strategy is eliminated, the model reduces averagely by 4.95\% \emph{Acc.} and 6.36\% \emph{F1}. They demonstrate generating adversarial target domain samples improves model robustness and implementing contrastive training effectively overcomes the inter-domain discrepancy.

\subsection{Case Study}
To further explore how COOL extract comprehensive knowledge to enhance prompt learning in FND, we illustrate a case in CoAID in Figure~\ref{fig: case}. The news reports the index case of coronavirus in China. For every linked entity, we retrieve both structured and unstructured knowledge from heterogeneous external sources. For instance, “first case” mentioned in news is linked to entity “Index case”, whose unstructured description is retrieved from knowledge corpus, and structured knowledge are filtered out from knowledge graph. The retrieved knowledge of all entities is then fed into the signed correlation-aware attention to extract knowledge that are positively or negatively correlated with news. Specifically, as shown in the heatmap, N-$\mathrm{E^{+}}$ attention focus more on knowledge from entities that are highly positively correlated with news, such as “Coronavirus” and “Index case”, while N-$\mathrm{E^{-}}$ attention pay more attention to knowledge from entities that are not very (i.e., negatively) correlated with news, like “Human sexual activity”. Intuitively, the extracted both positively and negatively correlated knowledge contribute to the authenticity judgement. The case study shows COOL can extract comprehensive and important knowledge for FND.

\section{Conclusions}
In this paper, we propose COOL, which extracts comprehensive knowledge from heterogeneous external sources and incorporates knowledge into hybrid prompt learning to verify news authenticity in domain adaptive few-shot scenario. The method is equipped with good expressiveness because: (i) we extract comprehensive knowledge that either positively or negatively correlate with news from both structured relational knowledge and unstructured description knowledge; (ii) we adopt hybrid prompt template which incorporates comprehensive knowledge freely by learned soft prompt and guides PLM in FND task by hand-crafted hard prompt; (iii) adversarial contrastive training is implemented to robustly model the domain-invariant news-knowledge interaction pattern. The extensive experiments on real-world datasets validate the effectiveness of COOL in domain adaptive few-shot FND and its capacity in incorporating comprehensive knowledge into prompt learning framework.

\section*{Limitations}
Our work insists in injecting comprehensive knowledge into prompt learning for domain adaptive few-shot FND. However, in order to retrieve up-to-date external knowledge for FND in emerging domains, our model crawls Wikidata to obtain structured entity neighbors and unstructured entity descriptions, which can be time-consuming in pre-processing stage. Also, we find that the existing entity linking methods may overlook important news entities in some cases, which is a bottleneck for providing comprehensive knowledge information. Additionally, despite investigating prompt learning in domain adaptive FND, we do not discuss cross-domain adaptation with several source domains to improve detection performance in target domain. 

\bibliography{acl_arxiv_version}

\begin{thebibliography}{44}
\providecommand{\natexlab}[1]{#1}

\bibitem[{Bahad et~al.(2019)Bahad, Saxena, and Kamal}]{bahad2019fake}
Pritika Bahad, Preeti Saxena, and Raj Kamal. 2019.
\newblock Fake news detection using bi-directional lstm-recurrent neural network.
\newblock \emph{Procedia Computer Science}, 165:74--82.

\bibitem[{Bai et~al.(2024)Bai, Zhang, Zhou, Huang, Luan, Wang, and Chen}]{bai2024prompt}
Shuanghao Bai, Min Zhang, Wanqi Zhou, Siteng Huang, Zhirong Luan, Donglin Wang, and Badong Chen. 2024.
\newblock Prompt-based distribution alignment for unsupervised domain adaptation.
\newblock In \emph{Proceedings of the AAAI Conference on Artificial Intelligence}, volume~38, pages 729--737.

\bibitem[{Chen(2015)}]{chen2015convolutional}
Yahui Chen. 2015.
\newblock Convolutional neural network for sentence classification.
\newblock Master's thesis, University of Waterloo.

\bibitem[{Cui and Lee(2020)}]{cui2020coaid}
Limeng Cui and Dongwon Lee. 2020.
\newblock Coaid: Covid-19 healthcare misinformation dataset.
\newblock \emph{arXiv preprint arXiv:2006.00885}.

\bibitem[{Dun et~al.(2021)Dun, Tu, Chen, Hou, and Yuan}]{dun2021kan}
Yaqian Dun, Kefei Tu, Chen Chen, Chunyan Hou, and Xiaojie Yuan. 2021.
\newblock Kan: Knowledge-aware attention network for fake news detection.
\newblock In \emph{Proceedings of the AAAI conference on artificial intelligence}, volume~35, pages 81--89.

\bibitem[{Ferragina and Scaiella(2010)}]{ferragina2010tagme}
Paolo Ferragina and Ugo Scaiella. 2010.
\newblock Tagme: on-the-fly annotation of short text fragments (by wikipedia entities).
\newblock In \emph{Proceedings of the 19th ACM international conference on Information and knowledge management}, pages 1625--1628.

\bibitem[{Ge et~al.(2023)Ge, Huang, Xie, Lai, Song, Li, and Huang}]{ge2023domain}
Chunjiang Ge, Rui Huang, Mixue Xie, Zihang Lai, Shiji Song, Shuang Li, and Gao Huang. 2023.
\newblock Domain adaptation via prompt learning.
\newblock \emph{IEEE Transactions on Neural Networks and Learning Systems}.

\bibitem[{Guo et~al.(2023)Guo, Diefenbach, Gourru, and Gravier}]{guo2023wikidata}
Kunpeng Guo, Dennis Diefenbach, Antoine Gourru, and Christophe Gravier. 2023.
\newblock Wikidata as a seed for web extraction.
\newblock In \emph{Proceedings of the ACM Web Conference 2023}, pages 2402--2411.

\bibitem[{Guo et~al.(2024)Guo, Lu, Yu, Nguyen, and Yin}]{10.1145/3589334.3645337}
Lei Guo, Ziang Lu, Junliang Yu, Quoc Viet~Hung Nguyen, and Hongzhi Yin. 2024.
\newblock \href {https://doi.org/10.1145/3589334.3645337} {Prompt-enhanced federated content representation learning for cross-domain recommendation}.
\newblock In \emph{Proceedings of the ACM on Web Conference 2024}, WWW '24, page 3139–3149, New York, NY, USA. Association for Computing Machinery.

\bibitem[{Hospedales et~al.(2021)Hospedales, Antoniou, Micaelli, and Storkey}]{hospedales2021meta}
Timothy Hospedales, Antreas Antoniou, Paul Micaelli, and Amos Storkey. 2021.
\newblock Meta-learning in neural networks: A survey.
\newblock \emph{IEEE transactions on pattern analysis and machine intelligence}, 44(9):5149--5169.

\bibitem[{Hu et~al.(2021)Hu, Yang, Zhang, Zhong, Tang, Shi, Duan, and Zhou}]{hu2021compare}
Linmei Hu, Tianchi Yang, Luhao Zhang, Wanjun Zhong, Duyu Tang, Chuan Shi, Nan Duan, and Ming Zhou. 2021.
\newblock Compare to the knowledge: Graph neural fake news detection with external knowledge.
\newblock In \emph{Proceedings of the 59th Annual Meeting of the Association for Computational Linguistics and the 11th International Joint Conference on Natural Language Processing (Volume 1: Long Papers)}, pages 754--763.

\bibitem[{Jiang et~al.(2022)Jiang, Liu, Zhao, Sun, and Zhang}]{jiang2022fake}
Gongyao Jiang, Shuang Liu, Yu~Zhao, Yueheng Sun, and Meishan Zhang. 2022.
\newblock Fake news detection via knowledgeable prompt learning.
\newblock \emph{Information Processing \& Management}, 59(5):103029.

\bibitem[{Kaliyar et~al.(2021)Kaliyar, Goswami, and Narang}]{kaliyar2021fakebert}
Rohit~Kumar Kaliyar, Anurag Goswami, and Pratik Narang. 2021.
\newblock Fakebert: Fake news detection in social media with a bert-based deep learning approach.
\newblock \emph{Multimedia tools and applications}, 80(8):11765--11788.

\bibitem[{Kim et~al.(2023)Kim, Park, Kwon, Jo, Thorne, and Choi}]{kim2023factkg}
Jiho Kim, Sungjin Park, Yeonsu Kwon, Yohan Jo, James Thorne, and Edward Choi. 2023.
\newblock Factkg: Fact verification via reasoning on knowledge graphs.
\newblock In \emph{Proceedings of the 61st Annual Meeting of the Association for Computational Linguistics (Volume 1: Long Papers)}, pages 16190--16206.

\bibitem[{Kingma and Ba(2014)}]{kingma2014adam}
Diederik~P Kingma and Jimmy Ba. 2014.
\newblock Adam: A method for stochastic optimization.
\newblock \emph{arXiv preprint arXiv:1412.6980}.

\bibitem[{Li et~al.(2023)Li, Wang, He, Zhang, and Liu}]{10.1145/3581783.3612501}
Jingqiu Li, Lanjun Wang, Jianlin He, Yongdong Zhang, and Anan Liu. 2023.
\newblock \href {https://doi.org/10.1145/3581783.3612501} {Improving rumor detection by class-based adversarial domain adaptation}.
\newblock In \emph{Proceedings of the 31st ACM International Conference on Multimedia}, MM '23, page 6634–6642, New York, NY, USA. Association for Computing Machinery.

\bibitem[{Li and Liang(2021)}]{li2021prefix}
Xiang~Lisa Li and Percy Liang. 2021.
\newblock Prefix-tuning: Optimizing continuous prompts for generation.
\newblock In \emph{Proceedings of the 59th Annual Meeting of the Association for Computational Linguistics and the 11th International Joint Conference on Natural Language Processing (Volume 1: Long Papers)}, pages 4582--4597.

\bibitem[{Lin et~al.(2022)Lin, Ma, Chen, Yang, Cheng, and Guang}]{lin2022detect}
Hongzhan Lin, Jing Ma, Liangliang Chen, Zhiwei Yang, Mingfei Cheng, and Chen Guang. 2022.
\newblock Detect rumors in microblog posts for low-resource domains via adversarial contrastive learning.
\newblock In \emph{Findings of the Association for Computational Linguistics: NAACL 2022}, pages 2543--2556.

\bibitem[{Lin et~al.(2023)Lin, Yi, Ma, Jiang, Luo, Shi, and Liu}]{lin2023zero}
Hongzhan Lin, Pengyao Yi, Jing Ma, Haiyun Jiang, Ziyang Luo, Shuming Shi, and Ruifang Liu. 2023.
\newblock Zero-shot rumor detection with propagation structure via prompt learning.
\newblock In \emph{Proceedings of the AAAI Conference on Artificial Intelligence}, volume~37, pages 5213--5221.

\bibitem[{Liu et~al.(2019)Liu, Ott, Goyal, Du, Joshi, Chen, Levy, Lewis, Zettlemoyer, and Stoyanov}]{liu2019roberta}
Yinhan Liu, Myle Ott, Naman Goyal, Jingfei Du, Mandar Joshi, Danqi Chen, Omer Levy, Mike Lewis, Luke Zettlemoyer, and Veselin Stoyanov. 2019.
\newblock Roberta: A robustly optimized bert pretraining approach.
\newblock \emph{arXiv preprint arXiv:1907.11692}.

\bibitem[{Ma et~al.(2023)Ma, Chen, Hou, and Yuan}]{ma2023kapalm}
Jing Ma, Chen Chen, Chunyan Hou, and Xiaojie Yuan. 2023.
\newblock Kapalm: Knowledge graph enhanced language models for fake news detection.
\newblock In \emph{Findings of the Association for Computational Linguistics: EMNLP 2023}, pages 3999--4009.

\bibitem[{Mehta et~al.(2022)Mehta, Pacheco, and Goldwasser}]{mehta2022tackling}
Nikhil Mehta, Mar{\'\i}a~Leonor Pacheco, and Dan Goldwasser. 2022.
\newblock Tackling fake news detection by continually improving social context representations using graph neural networks.
\newblock In \emph{Proceedings of the 60th Annual Meeting of the Association for Computational Linguistics (Volume 1: Long Papers)}, pages 1363--1380.

\bibitem[{Mosallanezhad et~al.(2022)Mosallanezhad, Karami, Shu, Mancenido, and Liu}]{mosallanezhad2022domain}
Ahmadreza Mosallanezhad, Mansooreh Karami, Kai Shu, Michelle~V Mancenido, and Huan Liu. 2022.
\newblock Domain adaptive fake news detection via reinforcement learning.
\newblock In \emph{Proceedings of the ACM Web Conference 2022}, pages 3632--3640.

\bibitem[{Mridha et~al.(2021)Mridha, Keya, Hamid, Monowar, and Rahman}]{mridha2021comprehensive}
Muhammad~F Mridha, Ashfia~Jannat Keya, Md~Abdul Hamid, Muhammad~Mostafa Monowar, and Md~Saifur Rahman. 2021.
\newblock A comprehensive review on fake news detection with deep learning.
\newblock \emph{IEEE access}, 9:156151--156170.

\bibitem[{Nan et~al.(2022)Nan, Wang, Zhu, Sheng, Shi, Cao, and Li}]{nan2022improving}
Qiong Nan, Danding Wang, Yongchun Zhu, Qiang Sheng, Yuhui Shi, Juan Cao, and Jintao Li. 2022.
\newblock Improving fake news detection of influential domain via domain-and instance-level transfer.
\newblock In \emph{Proceedings of the 29th International Conference on Computational Linguistics}, pages 2834--2848.

\bibitem[{Ouyang et~al.(2022)Ouyang, Wu, and Pan}]{ouyang2022asymmetrical}
Yi~Ouyang, Peng Wu, and Li~Pan. 2022.
\newblock Asymmetrical context-aware modulation for collaborative filtering recommendation.
\newblock In \emph{Proceedings of the 31st ACM International Conference on Information \& Knowledge Management}, pages 1595--1604.

\bibitem[{Pei et~al.(2023)Pei, Zhang, and Zhang}]{pei2023few}
Shichao Pei, Qiannan Zhang, and Xiangliang Zhang. 2023.
\newblock Few-shot low-resource knowledge graph completion with reinforced task generation.
\newblock In \emph{Findings of the Association for Computational Linguistics: ACL 2023}, pages 7252--7264.

\bibitem[{Popat et~al.(2017)Popat, Mukherjee, Str{\"o}tgen, and Weikum}]{popat2017truth}
Kashyap Popat, Subhabrata Mukherjee, Jannik Str{\"o}tgen, and Gerhard Weikum. 2017.
\newblock Where the truth lies: Explaining the credibility of emerging claims on the web and social media.
\newblock In \emph{Proceedings of the 26th International Conference on World Wide Web Companion}, pages 1003--1012.

\bibitem[{Ran and Jia(2023)}]{ran2023unsupervised}
Hongyan Ran and Caiyan Jia. 2023.
\newblock Unsupervised cross-domain rumor detection with contrastive learning and cross-attention.
\newblock In \emph{Proceedings of the AAAI Conference on Artificial Intelligence}, volume~37, pages 13510--13518.

\bibitem[{Rozsa et~al.(2016)Rozsa, Rudd, and Boult}]{rozsa2016adversarial}
Andras Rozsa, Ethan~M Rudd, and Terrance~E Boult. 2016.
\newblock Adversarial diversity and hard positive generation.
\newblock In \emph{Proceedings of the IEEE conference on computer vision and pattern recognition workshops}, pages 25--32.

\bibitem[{Schick and Sch{\"u}tze(2021)}]{schick2021exploiting}
Timo Schick and Hinrich Sch{\"u}tze. 2021.
\newblock Exploiting cloze-questions for few-shot text classification and natural language inference.
\newblock In \emph{Proceedings of the 16th Conference of the European Chapter of the Association for Computational Linguistics: Main Volume}, pages 255--269.

\bibitem[{Shu et~al.(2020)Shu, Mahudeswaran, Wang, Lee, and Liu}]{shu2020fakenewsnet}
Kai Shu, Deepak Mahudeswaran, Suhang Wang, Dongwon Lee, and Huan Liu. 2020.
\newblock Fakenewsnet: A data repository with news content, social context, and spatiotemporal information for studying fake news on social media.
\newblock \emph{Big data}, 8(3):171--188.

\bibitem[{Speer et~al.(2017)Speer, Chin, and Havasi}]{speer2017conceptnet}
Robyn Speer, Joshua Chin, and Catherine Havasi. 2017.
\newblock Conceptnet 5.5: An open multilingual graph of general knowledge.
\newblock In \emph{Proceedings of the AAAI conference on artificial intelligence}, volume~31.

\bibitem[{Suchanek et~al.(2007)Suchanek, Kasneci, and Weikum}]{suchanek2007yago}
Fabian~M Suchanek, Gjergji Kasneci, and Gerhard Weikum. 2007.
\newblock Yago: a core of semantic knowledge.
\newblock In \emph{Proceedings of the 16th international conference on World Wide Web}, pages 697--706.

\bibitem[{Sun et~al.(2023)Sun, Zhang, Ma, Xie, Liu, and Philip}]{sun2023inconsistent}
Mengzhu Sun, Xi~Zhang, Jianqiang Ma, Sihong Xie, Yazheng Liu, and S~Yu Philip. 2023.
\newblock Inconsistent matters: A knowledge-guided dual-consistency network for multi-modal rumor detection.
\newblock \emph{IEEE Transactions on Knowledge and Data Engineering}.

\bibitem[{Tseng et~al.(2022)Tseng, Yang, Wang, and Peng}]{tseng2022kahan}
Yu-Wun Tseng, Hui-Kuo Yang, Wei-Yao Wang, and Wen-Chih Peng. 2022.
\newblock Kahan: knowledge-aware hierarchical attention network for fake news detection on social media.
\newblock In \emph{Companion Proceedings of the Web Conference 2022}, pages 868--875.

\bibitem[{Vaswani et~al.(2017)Vaswani, Shazeer, Parmar, Uszkoreit, Jones, Gomez, Kaiser, and Polosukhin}]{vaswani2017attention}
Ashish Vaswani, Noam Shazeer, Niki Parmar, Jakob Uszkoreit, Llion Jones, Aidan~N Gomez, {\L}ukasz Kaiser, and Illia Polosukhin. 2017.
\newblock Attention is all you need.
\newblock \emph{Advances in neural information processing systems}, 30.

\bibitem[{Wu et~al.(2023)Wu, Li, Deng, Xiong, and Hooi}]{10.1145/3583780.3615015}
Jiaying Wu, Shen Li, Ailin Deng, Miao Xiong, and Bryan Hooi. 2023.
\newblock \href {https://doi.org/10.1145/3583780.3615015} {Prompt-and-align: Prompt-based social alignment for few-shot fake news detection}.
\newblock In \emph{Proceedings of the 32nd ACM International Conference on Information and Knowledge Management}, CIKM '23, page 2726–2736, New York, NY, USA. Association for Computing Machinery.

\bibitem[{Xiao et~al.(2024)Xiao, Zhang, Shi, Wang, Naseem, and Hu}]{10.1145/3589334.3645468}
Liang Xiao, Qi~Zhang, Chongyang Shi, Shoujin Wang, Usman Naseem, and Liang Hu. 2024.
\newblock \href {https://doi.org/10.1145/3589334.3645468} {Msynfd: Multi-hop syntax aware fake news detection}.
\newblock In \emph{Proceedings of the ACM on Web Conference 2024}, WWW '24, page 4128–4137, New York, NY, USA. Association for Computing Machinery.

\bibitem[{Yue et~al.(2022)Yue, Zeng, Kou, Shang, and Wang}]{yue2022contrastive}
Zhenrui Yue, Huimin Zeng, Ziyi Kou, Lanyu Shang, and Dong Wang. 2022.
\newblock Contrastive domain adaptation for early misinformation detection: A case study on covid-19.
\newblock In \emph{Proceedings of the 31st ACM International Conference on Information \& Knowledge Management}, pages 2423--2433.

\bibitem[{Yue et~al.(2023)Yue, Zeng, Zhang, Shang, and Wang}]{yue2023metaadapt}
Zhenrui Yue, Huimin Zeng, Yang Zhang, Lanyu Shang, and Dong Wang. 2023.
\newblock Metaadapt: Domain adaptive few-shot misinformation detection via meta learning.
\newblock In \emph{Proceedings of the 61st Annual Meeting of the Association for Computational Linguistics (Volume 1: Long Papers)}, pages 5223--5239.

\bibitem[{Zhang et~al.(2024)Zhang, Zhang, Zhou, Huang, and Li}]{zhang2024reinforced}
Litian Zhang, Xiaoming Zhang, Ziyi Zhou, Feiran Huang, and Chaozhuo Li. 2024.
\newblock Reinforced adaptive knowledge learning for multimodal fake news detection.
\newblock In \emph{Proceedings of the AAAI Conference on Artificial Intelligence}, volume~38, pages 16777--16785.

\bibitem[{Zhao et~al.(2022)Zhao, Zheng, Zeng, He, Geng, Jiang, Wu, and Xu}]{zhao2022adpl}
Lulu Zhao, Fujia Zheng, Weihao Zeng, Keqing He, Ruotong Geng, Huixing Jiang, Wei Wu, and Weiran Xu. 2022.
\newblock Adpl: Adversarial prompt-based domain adaptation for dialogue summarization with knowledge disentanglement.
\newblock In \emph{Proceedings of the 45th International ACM SIGIR Conference on Research and Development in Information Retrieval}, pages 245--255.

\bibitem[{Zou et~al.(2023)Zou, Zhang, and Zhao}]{zou2023decker}
Anni Zou, Zhuosheng Zhang, and Hai Zhao. 2023.
\newblock Decker: Double check with heterogeneous knowledge for commonsense fact verification.
\newblock In \emph{Findings of the Association for Computational Linguistics: ACL 2023}, pages 11891--11904.

\end{thebibliography}

\appendix

\section{Experiment Setup Details}
\label{sec:appendix_setup}

\subsection{Datasets Statistics}
\label{sec:data_stat}
Three datasets are used to conduct domain-adaptive few-shot FND experiments, where Snopes \citep{popat2017truth} is a domain-agnostic dataset which is extracted from a fact-checking website \footnote{https://www.snopes.com/} providing various news articles and corresponding labels. Politifact \citep{shu2020fakenewsnet} is a political related dataset collected from another fact-checking website \footnote{https://www.politifact.com/} specialized for US political system. CoAID \citep{cui2020coaid} is a healthcare dataset containing COVID-19 related news on websites and social platforms. In our experiments, the domain-agnostic dataset Snopes is adopted as source domain dataset while the domain-specific datasets, Politifact and CoAID, are used as target domain datasets. We filter out the articles whose URL is no longer accessible. The statistical details of the datasets after prepossessing are summarized in Table~\ref{tab: statis}.

\begin{table}[ht]
\centering
\small
\caption{Statistics of the datasets}
\label{tab: statis}
\begin{tabular}{cccc} 
\hline
Datasets         & Snopes  & Politifact  & CoAID     \\ 
\hline
\# News          & 710     & 886         & 2807      \\
\# Real          & 430     & 517         & 2652      \\
\# Fake          & 280     & 369         & 155       \\
Avg. \# words    & 690     & 1361        & 78        \\
Avg. \# entities & 126     & 239         & 18        \\
\hline
\end{tabular}
\end{table}

\subsection{Baseline Methods}
\label{sec: baselines}
To evaluate the performance of our proposed model, we compare COOL with several groups of models to conduct domain adaptive few-shot FND experiments:

The first group of models is neural network-based models:

\textbf{TextCNN} \citep{chen2015convolutional}: This method uses convolutional neural networks with multiple filter widths to extract text features which are further fed into pooling layer and fully connected layer for classification.

\textbf{Bi-LSTM} \cite{bahad2019fake}: This method utilizes bi-directional long short-term memory which exploits text sequence from front-to-back and back-to-front and recurrent neural network for FND.

The second group of models is knowledge enhanced neural network-based models:

\textbf{KAN} \citep{dun2021kan}: This method proposes a knowledge-aware attention network for FND by extracting external knowledge from knowledge graph that are most relevant to news semantics with attention mechanism.

The third group of models is PLM-based models:

\textbf{FT} \citep{liu2019roberta}: This is the standard fine-tuning method built on top of RoBERTa by feeding [CLS] embedding into task-specific linear layers to predict news veracity.

\textbf{ACLR} \citep{lin2022detect}: This is a state-of-the-art domain adaptive FND method that adapts features learned from rich source domain to low-resource target domain by developing adversarial augmentation mechanism and supervised contrastive training paradigm.

\textbf{PET} \citep{schick2021exploiting}: This is a prompt learning method that provides task-related hand-crafted prompt to reformulate input as cloze-style phrases to help PLM understand the given task.

\textbf{Soft-PT} \citep{li2021prefix}: This method uses learnable tokens to optimize a sequence of continuous task-specific vectors for prompt tuning instead of discrete prompt which is constrained to real words embeddings.

\textbf{RPL} \citep{lin2023zero}: This is a state-of-the-art prompt learning-based method that generates adversarial augmentation examples and introduces a prototypical verbalizer paradigm with designed contrastive learning framework for detection task.

The fourth group of models is knowledge enhanced PLM-based models:

\textbf{KPL} \citep{jiang2022fake}: This is a strong baseline, which applies prompt learning to FND and incorporates knowledge features extracted from entity sequence into learnable prompts.

\subsection{Implementation Details}
\label{sec: imple}
We further detail our implementation as follows: The mini-batch Adaptive Moment Estimation (Adam) \citep{kingma2014adam} is adopted as the optimizer, which can adaptively adjust the learning rate during the training phase. We utilize Tagme \cite{ferragina2010tagme} as the entity linking method, while Wikidata \citep{guo2023wikidata} is used as external knowledge sources to crawl entity neighbors and entity descriptions. The hyper-parameter settings are as follows: training batch size is 16, hidden dimension of PLM is 768, the learning rate is 2e-5, the trade-off parameter $\alpha$ is 0.5, the temperature parameter $\tau$ is 0.1. For all baselines, the optimal hyper-parameter settings are determined either by our experiments or suggested by previous works to ensure the best performance. For fair comparisons, the base version of RoBERTa and Wikidata are used as PLM and external source for all needed methods, respectively, and the self-training and PLM ensemble for PET are not implemented, following previous work \citep{10.1145/3583780.3615015}. As we address the situations where no propagation structure can be obtained, and the adopted source dataset Snopes doesn't have the propagation structure, the propagation position modeling and the response ranking for RPL are not implemented. We add the cross-entropy loss into the training of RPL which is not included in the original paper, confirming the model stability in our experiments. All of our experiments are run on one single NVIDIA RTX A6000 GPU. The reported comparative results are averaged from ten implementations with randomly choiced seeds.

\section{Comparative Analysis}
\label{sec: comp}
Apart from the observation that COOL consistently outperforms baseline methods, there are more conclusions can be drawn from the comparative results shown in Table \ref{tab: comp}.

First, PLM-based methods generally outperform neural network-based methods on all experimental settings, which indicates the strong capacity of PLM in extracting semantic features of news to model fake news pattern. Among PLM-based methods, PET outperforms FT in most settings, which demonstrates the superiority of prompt learning over fine-tuning in domain adaptive scenario. However, Soft-PT generally performs worst in PLM-based methods, which indicates relying solely on randomly initialized soft prompts cannot effectively guide PLM in reasoning about task-related news authenticity.

Second, there are some findings in the comparison of knowledge enhanced models. As we can see in Table \ref{tab: comp}, different from many FND scenarios, knowledge enhanced neural network-based method performs generally worse than neural network-based methods in our experiments. This suggests that without non-trivial designs to overcome the inter-domain discrepancy, the news-knowledge interaction captured by knowledge enhanced neural network-based model is domain-specific which degenerates the model performance in emerging news domain. While for PLM-based models, KPL enhanced by knowledge outperforms PET suggests PLM’s potential in incorporating knowledge to boost detection performance.

Third, KPL generally performs the second-best results in the experiments on Covid. This suggests that introducing knowledge information is especially helpful for FND in emerging news domains since CoAID may be more related to recent real-world knowledge that are not contained in PLM pre-training corpus. Moreover, ACLR and RPL generally perform the second-best results in the experiments on Politifact, showing that the adversarial augmentation and contrastive training equipe them with strong domain adaptive learning capacity. 

\end{document}